\documentclass{article}

\usepackage{arxiv}

\usepackage[utf8]{inputenc} 
\usepackage[T1]{fontenc}    
\usepackage{hyperref}       
\usepackage{url}            
\usepackage{booktabs}       
\usepackage{amsfonts}       
\usepackage{nicefrac}       
\usepackage{microtype}      
\usepackage{lipsum}
\usepackage{graphicx}

\usepackage{tabularx} 
\usepackage{longtable}
\usepackage{multirow}
\usepackage{abstract}

\usepackage{authblk} 
\usepackage{hyperref}

\graphicspath{ {./images/} }

\title{A Component-Based Survey of Interactions between Large Language Models and Multi-Armed Bandits}

\author[1]{\textbf{Siguang Chen}}
\author[1,2]{\textbf{Chunli Lv}}
\author[1,2]{\textbf{Miao Xie}\textsuperscript{*}}

\affil[1]{College of Information and Electrical Engineering, China Agricultural University, Beijing 100083, China}
\affil[2]{Key Laboratory of Agricultural Machinery Monitoring and Big Data Application, Ministry of Agriculture and Rural Affairs, Beijing 100083, China}

\affil[*]{Corresponding author: \texttt{xiemiao@cau.edu.cn}}

\begin{document}
\maketitle
\begin{abstract}
Large language models (LLMs) have become powerful and widely used systems for language understanding and generation, while multi-armed bandit (MAB) algorithms provide a principled framework for adaptive decision-making under uncertainty. This survey explores the potential at the intersection of these two fields. As we know, it is the first survey to systematically review the bidirectional interaction between large language models and multi-armed bandits at the component level. We highlight the bidirectional benefits: MAB algorithms address critical LLM challenges, spanning from pre-training to retrieval-augmented generation (RAG) and personalization. Conversely, LLMs enhance MAB systems by redefining core components such as arm definition and environment modeling, thereby improving decision-making in sequential tasks. We analyze existing LLM-enhanced bandit systems and bandit-enhanced LLM systems, providing insights into their design, methodologies, and performance. Key challenges and representative findings are identified to help guide future research. An accompanying GitHub repository that indexes relevant literature is available at \url{https://github.com/bucky1119/Awesome-LLM-Bandit-Interaction}.
\end{abstract}


\section{Introduction}
Large language models (LLMs) have recently emerged as powerful general-purpose systems for language understanding, reasoning, and generation. Built upon transformer architectures and enabled by large-scale pretraining and post-training techniques, modern LLMs are no longer standalone natural language processing (NLP) tools but have evolved into modular components that can be invoked, combined, and scheduled within larger AI systems. This paradigm shift has led to their widespread adoption in applications such as code generation~\cite{li2025large}, intelligent assistants~\cite{dong2023towards}, scientific discovery~\cite{jo2023promise}, and decision support~\cite{surveylsm}, fundamentally reshaping the design of contemporary intelligent systems~\cite{Naveed2025llmsurvey}.

Reinforcement learning (RL) studies how agents learn to make sequential decisions through interaction with an environment. Among its many formulations, \emph{multi-armed bandits} (MABs) represent a classical and efficient setting that focuses on online decision-making under uncertainty through the \textit{exploration} (trying new actions) and \textit{exploitation} (choosing known rewarding actions) trade-off. See ~\cite{sutton2018reinforcement} for an introduction to various RL methods and ~\cite{lattimore2020bandit} for a more in-depth look into bandit algorithms. Compared to full Markov decision processes (MDPs), bandit models rely on immediate feedback rather than long-horizon state transitions, leading to simpler algorithmic structures, lower computational overhead, and stronger system controllability. Due to these properties, bandit frameworks have been widely adopted in practical systems such as recommendation~\cite{liu2018transferable}, online advertising~\cite{geng2021comparison}, adaptive experimentation~\cite{zhao2024adaptive,bouneffouf2020survey}.

Recent advances in LLMs have revealed a strong structural complementarity between LLMs and bandit frameworks. On the one hand, LLMs can enhance bandit algorithms by providing rich contextual representations~\cite{baheri2023llms}, prior knowledge~\cite{alamdari2024jump}, and reward predictions~\cite{park2024llm}. On the other hand, bandit methods offer LLM-based systems a principled mechanism for adaptive online decision-making~\cite{chen2024efficient}, enabling tasks such as model selection~\cite{dai2024cost}, tool invocation~\cite{chen2025learning}, generation control~\cite{puthiya2024sequential}, and inference strategy selection~\cite{atalar2025neural}. This synergy has driven increasing interest in combining LLMs and bandits across a wide range of interactive and decision-critical applications.

Despite this growing body of cross-domain research, a systematic and modular understanding of the LLM–bandit intersection remains lacking. Prior surveys, such as Mui et al.~\cite{surveybandit} on adaptive learning and Bouneffouf et al.~\cite{bouneffouf2020survey} on bandit applications, primarily focus on general-purpose domains like recommendation or ad placement, and do not consider the emerging structure of LLM systems. Likewise, surveys on reinforcement learning in NLP~\cite{surveyreinforcement, luketina2019survey} and sequence models~\cite{surveylsm} do not focus on the specific intersection between bandit algorithms and large language models.
Concurrent with our work, Bouneffouf et al. ~\cite{bouneffouf2025surveymultiarmedbanditsmeet} also surveyed the interplay between bandit algorithms and large language models. While their work provides a conceptual overview of mutual enhancement between the two paradigms, our review focuses on a system-level component perspective that systematically maps interactions between LLM modules and bandit algorithmic components.

In this survey, we adopt a \emph{component-based perspective} on the interactions between large language models and multi-armed bandits. We organize existing work into two complementary categories: \emph{bandit-enhanced LLM systems}, where bandit algorithms improve the online decision-making and control of LLM pipelines, and \emph{LLM-enhanced bandit frameworks}, where LLMs augment key components of traditional bandits. This perspective enables a unified technical mapping of existing methods, clarifies common design patterns, and highlights open challenges at the intersection of large language models and multi-armed bandits.

Our main contributions are summarized as follows.

\begin{itemize}
  \item \textbf{Systematic Component-Level Survey of the Bandit--LLM Intersection:}
  To the best of our knowledge, this survey represents the first systematic review that analyzes the interaction between multi-armed bandit algorithms and large language models from a component-based and algorithmic perspective.

  \item \textbf{Unified Component-Based Taxonomic Framework:}
  We introduce a structured component-based taxonomy for both LLM systems and bandit systems, enabling a unified technical lens to examine how bandit mechanisms enhance LLM pipelines and how LLMs reshape core bandit components.

    \item \textbf{Identification of Challenges and Research Opportunities:}
  This survey critically examines existing cross-domain methods, synthesizes key technical and conceptual challenges, and outlines promising research directions that can guide future work at the intersection of bandit algorithms and large language models.
\end{itemize}

The organization of this article is as follows. Section~\ref{sec:background} introduces the background information about large language models and multi-armed bandits. Section~\ref{sec:survey-methodology} describes the survey methodology adopted in this work. A component-based taxonomic framework for both LLM systems and bandit systems is presented in Section~\ref{sec:taxonomy}. We then review representative studies on the bidirectional integration between LLMs and bandit algorithms in Sections~\ref{sec:bandit_based_enhancements_for_llm_systems} and~\ref{sec:llm-based_enhancements_for_bandit_systems}. Finally, we discuss challenges and future opportunities in Section~\ref{sec:challenges_and_future_opportunities} and conclude in Section~\ref{sec:conclusion}.

\section{Background}
\label{sec:background}
In this section, we present the relevant preliminary background to facilitate readers’ comprehension of large language models (LLMs) and bandit algorithms. For in-depth information, we recommend consulting the original works.

\subsection{Large Language Models}

Large language models (LLMs) are a class of neural language models designed to learn the statistical structure of natural language at scale. At their core, LLMs model the probability distribution of token sequences, enabling them to predict and generate text conditioned on a given context. By learning from large corpora of human-written text, LLMs acquire rich representations of linguistic patterns, factual knowledge, and semantic relationships. Comprehensive introductions to language modeling can be found in existing surveys such as~\cite{zhao2023llmsurvey,minaee2024llmsurvey,Naveed2025llmsurvey}.

Modern LLMs are predominantly built upon the transformer architecture~\cite{vaswani2017attention}, which employs self-attention mechanisms to capture long-range dependencies in text. A common training paradigm consists of large-scale self-supervised pre-training on unlabeled text data, followed by task-specific fine-tuning or alignment procedures to adapt the model to downstream applications~\cite{Naveed2025llmsurvey}. Depending on architectural design, pretrained language models can be categorized into encoder-only~\cite{devlin2019bert}, decoder-only~\cite{radford2019language}, and encoder–decoder models~\cite{raffel2020exploring}, each suited to different types of language understanding and generation tasks.

LLMs can be viewed as scaled-up extensions of earlier pretrained language models, characterized by substantially increased model capacity and training data volume. Representative examples include GPT-3~\cite{brown2020language}, PaLM~\cite{chowdhery2023palm}, and LLaMA~\cite{touvron2023llama}. Empirical studies on neural scaling laws suggest that increasing model size, data scale, and computational budget can lead to consistent performance improvements across a wide range of tasks~\cite{kaplan2020scaling,hoffmann2022training}. Moreover, LLMs have been observed to exhibit emergent capabilities—such as in-context learning, instruction following, and multi-step reasoning—that are not explicitly programmed but arise as a result of large-scale training~\cite{openai2023gpt4}.

Owing to these capabilities, LLMs have demonstrated strong performance not only in traditional natural language processing tasks, but also in a growing set of applications including information retrieval~\cite{zhang2024large}, code generation~\cite{li2025large}, science discovery~\cite{jo2023promise}, personalized digital assistants~\cite{dong2023towards}, education~\cite{guizani2025systematic}, finance~\cite{dong2025large}, and healthcare~\cite{iqbal2025impact}. Their ability to reason over unstructured text, integrate external knowledge, and interact with tools has further motivated research into system-level integration strategies, where LLMs are deployed as components within larger decision-making pipelines.

In this survey, rather than focusing on model architectures or training techniques, we adopt a system-level perspective that decomposes LLM-based systems into functional components. This abstraction enables a clearer analysis of how decision-making mechanisms—such as bandit algorithms—can be integrated with LLMs to improve learning efficiency, adaptability, and robustness. Detailed component definitions and taxonomy are presented in Section~\ref{llm taxonomy}.

\subsection{Multi-Armed Bandit}

A multi-armed bandit (MAB) problem models a sequential decision-making setting in which an agent repeatedly selects an action (arm) from a predefined set and observes stochastic feedback. See the literature~\cite{slivkins2019introduction,lattimore2020bandit} for a detailed paradigm of bandit algorithms and a more comprehensive overview of various algorithms. At each interaction, the agent balances exploration of uncertain actions and exploitation of actions with high expected reward, with the goal of optimizing long-term performance, typically measured by cumulative reward or regret.

Research in MAB has evolved from \textit{stochastic bandits}, where reward distributions are fixed but unknown, to \textit{contextual bandits}, where actions depend on contextual information~\cite{li2010contextual,perchet2013multi}. Further advancements include \textit{semi-parametric bandits}~\cite{peng2019practical,choi2023semi}, which integrate parametric models to estimate rewards, and \textit{neural bandits}, which leverage deep learning for breaking linear assumptions in large-scale contexts~\cite{zhou2020neural,chen2022interconnected}. Additionally, \textit{non-stationary bandits} address environments where reward distributions change over time~\cite{besbes2014stochastic}, while \textit{adversarial bandits} account for adversarial changes in the reward structure~\cite{auer2002nonstochastic}. Recent innovations have introduced the concept of \textit{meta-bandits}, exemplified by systems like \textit{AutoBandit}, which dynamically adapt and optimize bandit algorithms in real-time decision-making environments~\cite{xie2021autobandit}. These advancements have made MABs critical in applications such as \textit{online recommendations}, \textit{advertisement placement}, \textit{adaptive learning systems}, and even \textit{machine learning algorithms}~\cite{bouneffouf2020survey}. Despite their differences, these variants primarily differ in how they model rewards, represent actions, incorporate contextual information, and handle uncertainty over time.

As a result, MABs have been widely applied to practical decision-making problems such as online recommendation, advertisement placement, and adaptive learning systems~\cite{bouneffouf2020survey}, where decision pipelines often involve multiple interacting components. Motivated by this diversity of modeling assumptions and application requirements, bandit formulations vary depending on assumptions about the reward generation process, the availability of contextual information, and the structure of the action space. Accordingly, we adopt a system-level perspective and defer a detailed component-wise decomposition of bandit systems to the taxonomy presented in Section~\ref{subsec:bandit_component_taxonomy}.

\section{Survey Methodology}
\label{sec:survey-methodology}

To ensure a comprehensive and objective overview, this study follows a systematic review methodology~\cite{aromataris2014systematic,pollock2018systematic}. We conducted an extensive search across major scholarly databases using approximately $30$ key terms spanning the intersection of Multi-Armed Bandits (MABs) and Large Language Models (LLMs). Our initial search yielded over $300$ candidate papers. Through a rigorous manual screening process—focusing on the technical integration of bandit mechanisms within LLM workflows—we refined the selection to over $100$ core papers. To facilitate community research and ensure reproducibility, we have curated an open-source repository that indexes these works according to our proposed taxonomy. The complete list of hierarchical search queries and the detailed selected papers  are documented in the repository to support further systematic updates. This resource is available at \url{https://github.com/bucky1119/Awesome-LLM-Bandit-Interaction} and will be continuously maintained to reflect emerging advancements.

\section{Taxonomic Framework}
\label{sec:taxonomy}

\subsection{Component-Based Taxonomy of LLM Systems}
\label{llm taxonomy}

Following recent surveys on large language models (LLMs)~\cite{zhao2023llmsurvey,minaee2024llmsurvey,Naveed2025llmsurvey}, the lifecycle of LLMs can be systematically divided into two major stages: the \emph{building stage} and the \emph{augment stage}. While the building and augmentation stages describe the \emph{lifecycle} of LLM development and deployment, our taxonomy does not categorize methods by stages themselves, but rather by the \emph{functional system components} that operate within or across these stages.

\textit{\textbf{Building Stage}.} The goal of the building stage is to train a general-purpose foundation model that encodes broad world knowledge and strong language understanding capabilities. This stage typically consists of two sub-stages: pre-training and post-training~\cite{zhao2023llmsurvey}.

During the pre-training stage, LLMs are trained on large-scale corpora using self-supervised objectives. This process involves data collection and cleaning, tokenization, model architecture design (e.g., transformer-based architectures), positional encoding schemes, and large-scale optimization~\cite{Naveed2025llmsurvey}. Through pre-training, the model acquires general linguistic representations and extensive world knowledge.

The post-training stage further shapes the model’s capabilities and behavior to better align with downstream usage requirements. This stage commonly includes fine-tuning techniques such as instruction tuning, as well as alignment methods (e.g., reinforcement learning from human feedback, RLHF) that aim to align the model’s outputs with human preferences, values, and safety constraints~\cite{zhang2023instructionsurvey}. Together, the pre-training and post-training stages yield a general-purpose LLM that can serve as the backbone for a wide range of applications.

\small
\begin{table}[h]
  \caption{Component-Based Taxonomy of LLM Systems}
  \label{tab:llm_taxonomy}
  \setlength{\tabcolsep}{3pt}
  \renewcommand{\arraystretch}{1.10}
  \begin{tabularx}{\linewidth}{@{}p{3.2cm} X@{}}
    \toprule
    \textbf{LLM Component} & \textbf{Description} \\
    \midrule
    Pre-training &
    Large-scale self-supervised training on broad corpora to learn general linguistic representations and parametric world knowledge. \\
    \addlinespace[2pt]

    Fine-tuning &
    Supervised or instruction-based post-training that adapts a foundation model to downstream tasks or domains. \\
    \addlinespace[2pt]

    Alignment &
    Post-training procedures that steer model behavior toward human preferences and safety constraints using feedback signals (e.g., preference data or reward models). \\
    \addlinespace[2pt]

    Prompt Design and Selection &
    Designing and selecting prompts (including templates and exemplars) to specify tasks and control behavior at inference time without changing model parameters. \\
    \addlinespace[2pt]

    Tool and Function Calling &
    Mechanisms that enable LLMs to decide when and how to invoke external tools, APIs, or functions and integrate their outputs into generation. \\
    \addlinespace[2pt]

    Context Understanding &
    Structuring, filtering, and interpreting provided context (e.g., user intent, dialogue state, long context) to support coherent and relevant conditional generation. \\
    \addlinespace[2pt]

    Retrieval-Augmented Generation (RAG) &
    Retrieving external information and conditioning generation on retrieved evidence to complement parametric memory and improve factual grounding. \\
    \addlinespace[2pt]

    Inference Optimization &
    System-level strategies that reduce latency and cost (e.g., caching, batching, routing, speculative decoding) while maintaining output quality during deployment. \\
    \addlinespace[2pt]

    Decoding Strategies &
    Token-level generation procedures (e.g., sampling, beam search, reranking) that trade off diversity, stability, and quality. \\
    \addlinespace[2pt]

    Adaptation and Personalization &
    Mechanisms that tailor model behavior to users, tasks, or environments over time, enabling adaptive and personalized interactions. \\
    \bottomrule
  \end{tabularx}
\end{table}

\textbf{\textit{Augmentation Stage. }} Once an LLM has been trained, it can be directly applied to various tasks via prompting. However, in practical deployments, LLMs often exhibit notable limitations, including hallucinations, lack of persistent memory, large computational overhead, inherently stochastic generation behavior, and outdated or incomplete knowledge~\cite{minaee2024llmsurvey}. To address these issues and fully exploit the potential of LLMs, researchers increasingly rely on external, system-level enhancement mechanisms~\cite{Miao2025LLMEfficientsurvey}, many of which operate at inference time without (or with minimal) modification of model parameters. We refer to this phase as the \emph{augmentation stage}~\cite{minaee2024llmsurvey}. The augment stage encompasses a variety of techniques and system components designed to improve usability, robustness, and task performance. Representative examples include prompt design and selection, tool and function calling, retrieval-augmented generation (RAG), adaptation and personalization mechanisms~\cite{liu2025personalizedsurvey}, as well as inference optimization strategies~\cite{Plaat2025LLMReasoningSurvey}.

From a \textbf{\textit{systematic component-based breakdown}} perspective, we abstract LLM systems as compositions of decision-relevant functional modules that may reside at different stages of the LLM lifecycle. Considering that bandit algorithms—viewed as a minimal modeling paradigm within reinforcement learning for sequential decision-making under uncertainty—are naturally suited for characterizing and optimizing such decision-related modules (e.g., prompt selection, tool invocation, and context management), we adopt this perspective to structure our taxonomy.

As illustrated in table~\ref{tab:llm_taxonomy}, we identify a set of core components that together constitute an LLM system, including \textbf{pre-training}, \textbf{fine-tuning}, \textbf{alignment}, \textbf{prompt design and selection}, \textbf{tool and function calling}, \textbf{context understanding}, \textbf{retrieval-augmented generation (RAG)}, \textbf{inference optimization}, \textbf{decoding strategies}, and \textbf{adaptation and personalization}. While these components are instantiated at different stages of the LLM lifecycle, they are grouped based on their functional roles in the overall system rather than their temporal order.

For conceptual clarity, these components can be viewed as belonging to several high-level functional categories: training-related components (pre-training, fine-tuning, and alignment), input control mechanisms (prompt design and selection, and tool and function calling), context augmentation modules (context understanding and retrieval-augmented generation), inference and generation processes (inference optimization and decoding strategies), and output adaptation strategies (adaptation and personalization). However, to maintain consistency with the bandit taxonomy and to facilitate component-wise analysis in subsequent sections, we treat individual components as the primary units of organization. See Section~\ref{sec:bandit_based_enhancements_for_llm_systems} for detailed component-level discussions and representative bandit-based enhancement methods.

\subsection{Component-Based Taxonomy of Bandit Systems}
\label{subsec:bandit_component_taxonomy}

A bandit algorithm addresses a sequential decision-making problem in which an agent repeatedly selects actions under uncertainty and updates its strategy based on observed feedback, with the objective of optimizing long-term performance such as cumulative reward or regret~\cite{lattimore2020bandit,auer2002finite}. Despite the diversity of existing bandit formulations and algorithms~\cite{slivkins2019introduction,lattimore2020bandit}, their core decision processes can be characterized by a shared system structure.

From a system perspective, a bandit algorithm can be viewed as a composition of functional components that together define how the decision problem is specified and solved. Specifically, these components include the optimization objective that guides learning, the definition of the action (arm) space, assumptions about the environment dynamics, the formulation of the reward signal, and the exploration-driven decision mechanism used to select actions over time. Each component captures a distinct aspect of the bandit decision pipeline, from problem modeling to action execution.

Based on this perspective, we adopt a component-based taxonomy to organize bandit systems according to their constituent modules rather than their specific algorithmic forms.

\small
\begin{table}[h]
  \caption{Component-Based Taxonomy of Bandit Systems}
  \label{tab:bandit_taxonomy}
  \setlength{\tabcolsep}{4pt}
  \renewcommand{\arraystretch}{1.10}
  \begin{tabularx}{\linewidth}{@{}p{3.5cm} X@{}}
    \toprule
    \textbf{Bandit Component} & \textbf{Description} \\
    \midrule
    Regret Minimization  Objective &
    Defines the learning goal as minimizing cumulative regret, i.e., the performance gap between the learner’s chosen actions and an oracle benchmark (typically the best fixed arm, or the best policy in contextual settings). \\
    \addlinespace[2pt]

    Arm Definition &
    Specifies what constitutes an action (arm) and its structure, ranging from a finite set of discrete arms to structured/continuous action spaces, including context-dependent actions in contextual bandits. \\
    \addlinespace[2pt]

    Environment &
    States assumptions on how rewards are generated over time, such as stochastic and stationary rewards, non-stationary or drifting rewards, or adversarial reward sequences, which determine what guarantees are achievable. \\
    \addlinespace[2pt]

    Reward Formulation &
    Defines the feedback signal and its type, such as scalar reward, binary click, cost-sensitive reward, or vector/multi-objective reward, as well as whether feedback is full-information, bandit (partial), delayed, or corrupted/noisy. \\
    \addlinespace[2pt]

    Sampling Strategy &
    Describes how the algorithm balances exploration and exploitation when collecting data, e.g., optimism (UCB-style indices), randomized exploration (e.g., $\epsilon$-greedy), posterior sampling (Thompson sampling), or elimination-based schemes. \\
    \addlinespace[2pt]

    Action Decision &
    Specifies the concrete decision rule executed each round given current estimates/posteriors and constraints, e.g., selecting $\arg\max$ of an index, sampling an arm from a learned distribution, or choosing a constrained/safe action in combinatorial or risk-aware settings. \\
    \bottomrule
  \end{tabularx}
\end{table}

Concretely, within this taxonomy, each component corresponds to a distinct modeling or decision-making function in a bandit system. The \textbf{regret minimization objective} specifies the performance criterion that guides algorithm design and theoretical analysis. The \textbf{arm definition} determines the action space over which decisions are made, ranging from simple discrete arms to structured or contextual action representations. The \textbf{environment} component captures assumptions about reward generation dynamics, such as stationarity, non-stationarity, or adversarial behavior. The \textbf{reward formulation} defines how feedback is observed, including issues of noise, delay, sparsity, or partial observability. The \textbf{sampling strategy} defines how uncertainty is modeled and exploited to construct action selection policies, such as optimism-based methods or posterior sampling. The \textbf{action decision} component corresponds to the execution step that selects a concrete action at each interaction based on the chosen strategy. 

Table~\ref{tab:bandit_taxonomy} summarizes the key components considered in this taxonomy. Detailed discussions of each bandit component and their corresponding LLM-based enhancement mechanisms are provided in Section~\ref{sec:llm-based_enhancements_for_bandit_systems}.

Having established this component-based decomposition, we are able to systematically map how bandit techniques enhance specific components of LLM systems and, conversely, how LLMs can improve traditional bandit algorithms. By dissecting both domains at the component level, this framework helps uncover previously hidden synergies, identify shared challenges, and provide clearer insights into their integration. Moreover, it enables evaluation across multiple dimensions, including research problems, methodologies, validation protocols, and datasets, thereby providing a structured foundation for understanding and advancing the intersection of MAB algorithms and LLMs.

\section{Bandit-Based Enhancements for LLM Systems}
\label{sec:bandit_based_enhancements_for_llm_systems}

This section provides a component-level review of how bandit algorithms are integrated into large language models to improve training, inference, and adaptation. Table~\ref{tab:bandit-for-llm} offers a consolidated overview of selected works discussed in this section.

\begin{table}[h]
\caption{Bandit-Based Enhancements for LLM Systems}
\label{tab:bandit-for-llm}

\centering
\scriptsize
\setlength{\tabcolsep}{3pt}
\renewcommand{\arraystretch}{1.1}

\begin{tabular}{p{2.4cm} p{5.6cm} p{5.6cm} p{1.8cm}}

\toprule
\textbf{LLM Component} &
\textbf{Research Problem} &
\textbf{Bandit-Enhanced Solutions} &
\textbf{References} \\
\midrule

\multirow{3}{2.4cm}{Pre-training} &
Static masking strategy selection &
Bandit-based dynamic masking pattern selection &
\cite{mukherjee2024pretraining} \\
\addlinespace[1pt]
& Static task sampling in multi-task pre-training &
Structured multi-task bandit formulation &
\cite{urteaga2022multi} \\
\addlinespace[1pt]
& Suboptimal trade-off between data quality and diversity &
Online domain-level sampling ratio optimization &
\cite{albalak2023efficient,zhang2024harnessing,ma2025actorcriticbasedonlinedata} \\

\midrule

\multirow{3}{2.4cm}{Fine-tuning} &
Reward overfitting in RL-based fine-tuning &
Adaptive training data reweighting and selection &
\cite{zhu2024dynamic,zhu2024iterative,zhou2024reflect,zhao2024sharp} \\
\addlinespace[1pt]
& Inefficient preference data collection &
Bandit-based preference and context selection &
\cite{duan2025chunks,liu2024sample,tajwar2024preference} \\
\addlinespace[1pt]
& Static budget allocation under non-stationarity &
Bandit-based adaptive dataset mixture optimization &
\cite{shin2025dynamixsft,xia2024llm,xia2024convergence} \\

\midrule

\multirow{3}{2.4cm}{Alignment} &
Costly and inefficient preference feedback collection &
Adaptive bandit-based preference query selection &
\cite{liu2024sampleefficient,mehta2025sample,lin2025activedpo} \\
\addlinespace[1pt]
& Non-stationary and myopic alignment dynamics &
Bandit-style exploration in online preference learning &
\cite{bai2025online,Calandriello2024Human,munos2024nash} \\
\addlinespace[1pt]
& Structured decision stages in alignment pipelines &
Bandit-guided structured alignment pipelines &
\cite{duan2025chunks} \\

\midrule

\multirow{4}{2.4cm}{Prompt Design\\and Selection} &
Budget-limited prompt evaluation &
Bandit-based best prompt identification &
\cite{shi2024best,shi2024efficient,ashizawa2025bandit} \\
\addlinespace[1pt]
& Implicit and costly prompt feedback &
Preference-based and dueling bandit optimization. &
\cite{lin2024prompt,wu2025llm} \\
\addlinespace[1pt]
& Suboptimal prompt composition &
Bandit-based example and trajectory selection &
\cite{wu2024prompt,rietz2025prompttuningdecisiontransformers,kong2025meta} \\
\addlinespace[1pt]
& Large-scale and privacy-constrained prompt spaces &
Offline, neural, and federated bandit optimization &
\cite{kiyohara2024prompt,kiyohara2025prompt,tan2024prompt,lin2024use,lu2025fedpobsampleefficientfederatedprompt} \\

\midrule

\multirow{3}{2.4cm}{Tool and\\Function Calling} &
Delayed and noisy feedback in multi-step tool use &
Execution-feedback-driven bandit tool selection &
\cite{zeng2025reinforcing,NEURIPS2024_c0f7ee19} \\
\addlinespace[1pt]
& Large and heterogeneous tool spaces &
Semantic-aware bandit-based tool generalization &
\cite{muller2025semantic,chen2025learning} \\
\addlinespace[1pt]
& Delayed feedback and credit assignment &
Bandit-informed agentic tool optimization &
\cite{jiang2025verltool,zhai2025agentevolver} \\

\midrule

\multirow{3}{2.4cm}{Context\\Understanding} &
Exploration and adaptation under sparse feedback &
Bandit-based in-context exploration &
\cite{krishnamurthy2024can,rahn2024controlling,dwaracherla2024efficient,tang2024code} \\
\addlinespace[1pt]
& Inefficient long-context utilization &
Adaptive bandit-guided context selection &
\cite{duan2025chunks,pan2025camab} \\
\addlinespace[1pt]
& Non-stationary inference-time decisions &
Contextual bandit-based online strategy selection. &
\cite{qin2024enhancing,nie2025evolve,poon2025multi,ramesh2025efficient} \\

\midrule

\multirow{3}{2.4cm}{Retrieval-Augmented\\Generation (RAG)} &
Adaptivity requirements in retrieval control across queries and budgets &
Bandit-based retrieval strategy selection &
\cite{tang2025mbaragbanditapproachadaptive,11044685,wang2024mragreinforcinglargelanguage} \\
\addlinespace[1pt]
& Heterogeneity and context dependence of retrieved evidence &
Bandit-driven knowledge source selection &
\cite{NEURIPS2024_0b8705a6,pan2025camab} \\
\addlinespace[1pt]
& Multi-objective and hierarchical RAG optimization. &
Hierarchical and multi-objective bandit optimization &
\cite{tang2025adapting,fu2024autoraghpautomaticonlinehyperparameter} \\

\midrule

\multirow{2}{2.4cm}{Inference\\Optimization} &
Static and inefficient inference model selection &
Adaptive bandit-based LLM routing &
\cite{li2025llmbanditcostefficientllm,panda2025adaptive,tongay2025dynamic,wei2025learningroutellmsbandit} \\
\addlinespace[1pt]
& High inference cost and resource waste &
Bandit-driven cache and resource optimization &
\cite{yang2025llm} \\

\midrule

\multirow{3}{2.4cm}{Decoding\\and Reranking} &
Limited generalization of static decoding and reranking heuristics &
Adaptive decoding configuration selection &
\cite{hou2025banditspec,liu2024speculative} \\
\addlinespace[1pt]
& Rich verification signals in speculative decoding &
Verification-augmented online decoding optimization &
\cite{liu2025not} \\
\addlinespace[1pt]
& Irreversible dependencies in token-level decoding &
Token-level bandit modeling &
\cite{shin2025tokenized} \\

\midrule

\multirow{3}{2.4cm}{Adaptation\\and Personalization} &
Heterogeneity and non-stationarity of user preferences &
Online user behavior exploration--exploitation &
\cite{chen2024online,goncc2023user,monea2024llms,moerchen2020personalizing} \\
\addlinespace[1pt]
& Performance--cost trade-offs in multi-task personalization &
Bandit-based personalized model selection and routing &
\cite{dai2024cost,li2025llmbanditcostefficientllm,wei2025learningroutellmsbandit} \\
\addlinespace[1pt]
& Inference-time personalization without retraining &
Bandit-driven inference-time preference adaptation &
\cite{shin2025tokenized,zhang2025exploration} \\

\bottomrule
\end{tabular}
\end{table}


\subsection{Pre-training}

Pre-training is the stage where an LLM acquires general-purpose language representations from massive corpora, and it is dominated by decisions about which data to sample and how to allocate computation. Traditional heuristics for data mixture and curriculum design are often static and manually tuned, which makes it hard to adapt to shifting data quality, domain distributions, and task requirements. Therefore, integrating MAB into pre-training offers a principled way to treat data selection, masking, and task scheduling as sequential decision problems under uncertainty, with feedback defined directly by pre-training progress and downstream performance~\cite{zhang2024harnessing, albalak2023efficient, mukherjee2024pretraining, urteaga2022multi}.

Recent work explores several complementary ways to couple MAB with LLM pre-training. 

One line of methods uses bandit-based policies to adapt training dynamics, for example by treating masking patterns as arms and updating their selection based on reward signals such as loss reduction; Mukherjee et al.~\cite{mukherjee2024pretraining} follow this idea and show that a bandit-driven dynamic masking scheme can reduce the number of training iterations and the need for extensive hyper-parameter search. 

A second line formulates multi-task pre-training as a structured bandit problem, where each task or task cluster is an arm and the bandit exploits correlations among them; Urteaga et al.~\cite{urteaga2022multi} implement this view through a decision transformer that predicts task-specific rewards and leverages task-level structure to improve data efficiency and generalization. 

A third line focuses on online data mixture and selection. Albalak et al.~\cite{albalak2023efficient} propose an online data mixing (ODM) algorithm that treats domain-level sampling ratios as bandit arms and adjusts them to maximize information gain, Zhang et al.~\cite{zhang2024harnessing} introduce Quad, which uses MAB to balance data quality and diversity in a scalable selection pipeline, and Ma et al.~\cite{ma2025actorcriticbasedonlinedata} extend this idea with AC-ODM, which models mixture optimization as an actor–critic process and uses gradient-alignment rewards to adapt domain sampling for better cross-domain generalization. 

Despite these advances, existing bandit-based pre-training methods still face limitations in reward design and granularity. 
Most approaches rely on short-horizon or proxy-based rewards (e.g., loss or perplexity) and operate at coarse domain or task levels, leaving open challenges in aligning bandit feedback with long-term generalization and in extending adaptive control to finer-grained data or training decisions.

\subsection{Fine-tuning}

Fine-tuning aims to adapt LLMs to specific domains and align them with human feedback, but RL-based fine-tuning introduces domain-specific challenges such as reward overfitting, unstable multi-step updates, and brittle generalization in low-resource settings, as observed in Myanmar dialect translation~\cite{thu2023rl}. Traditional supervised or RL pipelines rely on static datasets and manually tuned exploration schedules, complicating the balance between efficiency and robustness while risking myopic optimization toward short-term, noisy rewards~\cite{zhou2024reflect}. Therefore, recent work increasingly formulates fine-tuning as a sequential decision problem, where MAB-style allocation of data, rollouts, and compute can be used to stabilize learning and improve the feedback efficiency of LLM alignment.

Recent methods can be broadly grouped into three lines. 

First, curriculum and distribution-shaping approaches control the training signal by dynamically reweighting data and regularizing policy updates. Dynamic data mixing adjusts sampling weights according to dataset redundancy and correlation to optimize instruction tuning of Mixture-of-Experts models~\cite{zhu2024dynamic}, while Iterative Data Smoothing progressively refines soft labels to alleviate reward overfitting and improve reward convergence~\cite{zhu2024iterative}. Complementary work proposes a dual-model online RL framework in which a policy model and a reflection model cooperate to refine decisions from human feedback~\cite{zhou2024reflect}, and analyzes KL-regularization as a mechanism that can reduce sample complexity and improve policy efficiency in RLHF~\cite{zhao2024sharp}. Although not always framed explicitly as MAB, these approaches implicitly tackle an exploration–exploitation trade-off over data and rewards by adapting the effective training distribution. 

Second, bandit-based preference and data selection methods explicitly treat samples or context fragments as arms and optimize which feedback to collect. Duan et al.~\cite{duan2025chunks} introduce a chunk-sampling framework where a MAB mechanism selects context segments based on reward feedback and multi-round rollouts, using the resulting diverse responses to construct preference data and strengthen DPO training for long-context reasoning. Liu et al.~\cite{liu2024sample} model alignment as a contextual dueling bandit problem, using Thompson Sampling and an uncertainty-aware reward model to actively choose comparison pairs, thereby increasing the utility of each preference query and providing a unified banditized optimization view of online alignment. Tajwar et al.~\cite{tajwar2024preference} analyze online sampling and negative gradients in preference fine-tuning, and integrate bandit strategies into both data sampling and reward modeling to better exploit limited preference datasets. 

Third, bandit-driven model and dataset selection methods cast different LLMs or data sources as arms and dynamically allocate fine-tuning budgets in non-stationary environments. DynamixSFT formulates dataset mixing as a MAB problem and adaptively tunes the mixture during fine-tuning to improve the use of heterogeneous corpora~\cite{shin2025dynamixsft}. Xia et al.~\cite{xia2024llm} model online model selection as a non-stationary bandit, proposing TI-UCB to exploit the empirical pattern that reward first increases and then saturates with more fine-tuning steps, and thus to distribute limited budget across candidate LLMs more effectively. Building on this idea, Xia et al.~\cite{xia2024convergence} predict performance trends over training steps and combine UCB with change-detection mechanisms to decide which model to continue fine-tuning so as to maximize long-term return under strict resource constraints.

However, these approaches also have notable limitations. Dynamic data mixing may face scalability issues as data grow and can introduce additional noise into the training signal~\cite{zhu2024dynamic}, while Iterative Data Smoothing has not yet been extensively validated in more complex or highly non-stationary environments~\cite{zhu2024iterative}. Future work should broaden the applicability of bandit-enhanced fine-tuning to multilingual and multimodal settings, develop richer arm and reward abstractions that better capture long-horizon alignment objectives, and provide stronger theoretical guarantees for non-stationary and hierarchical bandit formulations that arise in large-scale LLM training.

\subsection{Alignment}
In LLM systems, the alignment component aims to steer model behavior toward human preferences, safety constraints, and task-specific objectives~\cite{bai2022training,wang2023aligning}. Beyond static supervised fine-tuning, alignment is increasingly framed as a sequential and interactive process, where models are repeatedly evaluated and adjusted based on feedback signals such as human judgments, preference comparisons, or implicit user satisfaction~\cite{Ouyang2022hf}. These feedback signals are often noisy, sparse, delayed, and costly to obtain, which poses significant challenges for efficiently selecting alignment actions and updating policies~\cite{shin2025tokenized}. From a decision-making perspective, alignment can be viewed as a process of repeatedly choosing alignment actions—such as preference queries, reward model updates, or policy adjustments—under uncertainty about their long-term effects~\cite{NEURIPS2023_a85b405e,pmlr-v235-munos24a}. By framing alignment as a bandit problem, LLM systems can adaptively balance these trade-offs and improve alignment outcomes under limited feedback budgets.

Recent work leverages multi-armed bandit (MAB) algorithms to improve alignment efficiency and robustness along several complementary lines.

A first line of work focuses on \emph{sample-efficient preference query and data selection}. In this setting, candidate preference comparisons or training examples are treated as bandit arms, and the learner adaptively selects which queries to present to human annotators under a constrained annotation budget. Several studies formulate preference annotation as an exploration--exploitation problem and use uncertainty-aware or confidence-based criteria to prioritize informative comparisons, significantly reducing the cost of human feedback while preserving alignment quality~\cite{liu2024sampleefficient,mehta2025sample}. More recent approaches further provide theoretically grounded selection rules for non-linear preference models. In particular, ActiveDPO derives uncertainty measures inspired by neural dueling bandits and actively selects preference data based on the gradients of the current LLM, enabling data selection strategies that are explicitly tailored to the model being aligned~\cite{lin2025activedpo}.

A second line of work views alignment as an \emph{online and iterative preference optimization process}, emphasizing the sequential and potentially non-stationary nature of human feedback. From this perspective, alignment is modeled as repeated interaction with a preference environment, where policies are continuously updated based on newly collected comparisons generated by the current model. Count-based and optimism-driven exploration strategies instantiate bandit-style decision rules in online preference optimization, encouraging systematic exploration of the prompt--response space during alignment~\cite{bai2025online}. Related formulations establish connections between online preference optimization, self-play, and game-theoretic learning dynamics, showing that stable aligned policies can be interpreted as equilibria of repeated preference learning processes~\cite{Calandriello2024Human,munos2024nash}.

Beyond direct policy updates, a third line of work explores \emph{structured alignment decisions} within the alignment pipeline using bandit abstractions. Instead of treating alignment solely as parameter optimization, these methods apply MAB algorithms to intermediate decision points that influence preference data construction. For example, recent work treats long-context chunks as bandit arms and uses MAB-guided sampling to adaptively select informative subsets of context for response generation, which are then used to construct high-quality preference pairs for DPO training~\cite{duan2025chunks}. This perspective highlights the flexibility of bandit formulations in capturing modular and structured alignment decisions beyond standard policy optimization.

Overall, framing alignment as a bandit problem provides a unified abstraction for handling uncertainty, feedback scarcity, and adaptivity in human-in-the-loop optimization. Nevertheless, existing bandit-based alignment methods face several challenges. Many formulations rely on simplified assumptions about preference feedback or implicit reward structures, while empirical studies show that alignment performance can be sensitive to reward scaling, regularization, and data distribution mismatch~\cite{liu2024sampleefficient,huang2025larger}. In addition, online alignment settings reveal a strong coupling between exploration strategies and policy optimization dynamics: although adaptive data generation improves coverage, it may also destabilize training if exploration is not properly regularized~\cite{Calandriello2024Human,munos2024nash}. 

These limitations also point to promising opportunities. Recent work suggests that bandit reasoning can be extended beyond isolated decisions to coordinate multiple alignment components, such as preference querying and structured input selection~\cite{duan2025chunks}. Moreover, advances in online preference optimization indicate that richer bandit formulations capable of handling non-stationary feedback may further improve alignment robustness~\cite{bai2025online,lin2025activedpo}.

\subsection{Prompt Design and Selection}

Prompt design and Selection aims to select instructions and in-context examples that elicit high-quality outputs from an LLM under strict query and computation budgets. The core challenge is that LLM performance is highly sensitive to prompt choices, while the model behaves as a black box and each query is costly. Traditional heuristic or manually designed prompts do not scale to large candidate spaces, struggle with complex example relationships and privacy constraints, and provide little principled support for trading off exploration and exploitation. Therefore, integrating MAB formulations provides a natural way to model prompt variants as arms, optimize under bandit feedback, and formalize budget-aware prompt search~\cite{yang2025multi}.

Recent work leverages MAB algorithms to structure prompt optimization along several complementary lines. 

A first line casts prompt selection as best-arm identification from a fixed candidate pool, seeking the most effective prompt under tight evaluation budgets; here, prompts or even prompt-optimization strategies are arms, and bandit procedures control which variants to evaluate, as in best-arm formulations for prompt pools and efficient variants that exploit prompt embeddings and clustering to scale to large candidate sets~\cite{shi2024best, shi2024efficient, ashizawa2025bandit}. 

A second line adopts preference-based and dueling bandits, where feedback is given via pairwise comparisons or implicit preferences instead of numerical scores; dueling frameworks use human feedback to adapt prompt choice and, more recently, dueling MAB with double Thompson sampling and prompt mutation to optimize prompts without relying on labeled data~\cite{lin2024prompt, wu2025llm}. 

A third line focuses on bandit-based selection and ordering of in-context examples or trajectory prompts, treating example subsets or trajectories as arms and optimizing them for in-context learning or multi-task reinforcement learning; representative studies optimize the composition and order of examples for better in-context learning, select trajectory-level prompts for decision transformers, and use meta-bandit schemes such as EXPO and EXPO-ES to tune task descriptions, meta-instructions, and example sets for sequential decision tasks~\cite{wu2024prompt, rietz2025prompttuningdecisiontransformers, kong2025meta}. 

A fourth line develops offline, federated, and neural bandit approaches that scale to large prompt spaces and privacy-sensitive settings: logged bandit data and clustering are used to reduce variance in gradient-based prompt tuning, off-policy bandit optimization (DSO) improves prompt policies for personalized recommendation tasks in large candidate spaces, neural bandits are combined with transformers or soft prompts to optimize black-box LLMs under query limits, and federated bandit algorithms such as FedPOB optimize prompts by sharing model parameters rather than raw data across parties~\cite{kiyohara2024prompt, kiyohara2025prompt, tan2024prompt, lin2024use, lu2025fedpobsampleefficientfederatedprompt}.

Despite these advances, existing approaches often assume a pre-generated prompt or strategy pool, which constrains exploration of genuinely novel prompts and limits adaptation to evolving tasks. Several methods also face practical issues such as high computation cost in large-scale sequential decision tasks, substantial communication overhead in federated prompt optimization, and misidentification risks in best-arm procedures over very large candidate sets~\cite{kong2025meta, lu2025fedpobsampleefficientfederatedprompt, shi2024efficient}. Future research should therefore develop more adaptive and generative bandit formulations that jointly expand and evaluate the prompt space, improve efficiency and robustness for large-scale deployments, and reduce reliance on dense human feedback and heavily engineered candidate pools.

\subsection{Tool and Function calling}
Tool and function calling enables LLM systems to interact with external APIs, databases, and software tools, extending their capabilities beyond pure text generation~\cite{plaat2025agentic}. At inference time, models must decide whether and how to invoke tools under partial or ambiguous context, where incorrect usage can incur nontrivial costs such as latency, execution failures, or degraded user experience~\cite{qu2025tool,Wu_2025_ICCV}. These decisions are inherently sequential and uncertain, as the utility of a tool invocation depends on task context, user intent, and downstream outcomes~\cite{chen2025learning}. Bandit-based formulations therefore offer a principled abstraction for modeling tool selection as an exploration--exploitation problem, enabling adaptive and cost-aware tool usage through interaction.

Recent work improves tool and function calling in LLM systems along several complementary lines. 

A first line models tool invocation as a multi-turn, error-aware reasoning process, showing that learning from execution feedback and assigning credit at finer temporal granularity can substantially improve robustness in multi-step tool-use scenarios~\cite{zeng2025reinforcing,NEURIPS2024_c0f7ee19}. 

A second line addresses scalability by incorporating semantic tool descriptions, contextual representations, or evolving abstractions, allowing tool selection policies to generalize across large and dynamic tool spaces~\cite{muller2025semantic,chen2025learning}. 

A third line adopts a system-level view, framing tool use as a long-horizon learning problem in agentic settings, where reinforcement learning optimizes tool-calling policies through interaction while exposing challenges such as delayed feedback and credit assignment~\cite{jiang2025verltool,zhai2025agentevolver}.

Despite these advances, tool calling remains challenging due to limited generalization across heterogeneous tools, difficulty in credit assignment under long-horizon interactions, and increasing uncertainty and cost in large tool ecosystems. These challenges also point to promising opportunities: integrating bandit-based decision-making with agentic reinforcement learning and rich contextual representations offers a scalable and adaptive framework for robust tool orchestration, positioning tool and function calling as a core sequential decision component in LLM-based systems.

\subsection{Context Understanding}

Contextual understanding in LLM-based decision systems aims to interpret rich, evolving input streams and translate them into adaptive exploration–exploitation behavior. Traditional supervised and reinforcement learning methods often struggle to scale in high-context regimes and to track non-stationary user preferences, especially when feedback is sparse or costly to obtain. Integrating MAB with LLMs is therefore valuable, as MAB provides principled uncertainty-aware exploration and feedback-efficient learning, while LLMs contribute semantic reasoning and in-context adaptation over complex textual or structured environments~\cite{krishnamurthy2024can, dwaracherla2024efficient, tang2024code}.

Recent work can be grouped into three main directions. 

First, a line of research studies bandit-driven in-context exploration and feedback-efficient learning. Krishnamurthy et al.~\cite{krishnamurthy2024can} use MAB benchmarks to probe how well LLMs perform in-context exploration, revealing when prompt-based adaptation suffices and when explicit bandit structure is needed. Building on this idea, Rahn et al.~\cite{rahn2024controlling} manipulate internal activations to steer exploratory behavior, effectively coupling bandit-style exploration objectives with mechanistic control of LLM representations. To reduce the cost of supervision, Dwaracherla et al.~\cite{dwaracherla2024efficient} propose a double Thompson sampling scheme that maintains competitive performance while minimizing feedback queries, and Tang et al.~\cite{tang2024code} cast code repair as a bandit problem where Thompson sampling over candidate patches guides LLM-based program edits. 

Second, several methods use bandit algorithms to select and attribute context in long or structured documents. Duan et al.~\cite{duan2025chunks} treat document chunks as arms in a LongMab-PO framework and use UCB-style selection to focus the LLM on the most informative combinations of text segments for long-context reasoning. Pan et al.~\cite{pan2025camab} similarly model each passage as an arm and apply adaptive sampling under a query budget to identify context most critical to answering, thereby making the LLM’s context dependencies explicit and more efficiently exploited. 

Third, a growing body of work uses contextual bandits to model user preferences, select among multiple LLMs, or guide inference-time decisions, including adversarial ones. Qin et al.~\cite{qin2024enhancing} introduce the MoRE framework, which combines explicit and implicit preferences in a multi-view reflection system to improve sequence modeling in recommendation tasks. Nie et al.~\cite{nie2025evolve} incorporate bandit algorithms at inference time so that the LLM can adaptively choose reasoning strategies based on contextual signals. Poon et al.~\cite{poon2025multi} formulate online model selection as a contextual bandit, using a greedy LinUCB policy to dynamically pick the most suitable LLM per step rather than modeling future context evolution. In a different direction, Ramesh et al.~\cite{ramesh2025efficient} design a MAB-based context-switching query procedure that incrementally elicits harmful outputs, demonstrating how bandit-guided querying can probe and circumvent LLM safety mechanisms.

Despite these advances, current approaches still face important limitations. Many exploration schemes are evaluated in simplified settings and may fail to generalize to highly dynamic, safety-critical environments with complex feedback channels. Bandit-guided context selection, activation control, and multi-LLM routing can introduce non-trivial computational overhead, especially when managing high-entropy policies or large search trees. Future work should therefore prioritize scalable exploration algorithms that remain robust under distribution shift, while designing more adaptive, feedback-efficient bandit frameworks that can operate with noisy, delayed, or partially observable signals~\cite{dwaracherla2024efficient, qin2024enhancing}.

\subsection{Retrieval-Augmented Generation (RAG)}

Retrieval-Augmented Generation (RAG) integrates external knowledge sources into LLM inference to improve factual consistency and interpretability. However, deciding when to retrieve, how many documents to fetch, and which strategy to use under varying query complexity, context requirements, and resource constraints remains a central challenge. Therefore, combining LLMs with MAB provides a principled way to cast retrieval control as online decision-making under uncertainty, balancing utility, latency, and computational cost.

A first line of work formulates retrieval strategy selection as a bandit problem to adapt retrieval intensity and prompt configuration to query difficulty and deployment conditions. MAB-RAG models candidate retrieval strategies as arms and uses online feedback to adjust retrieval behavior for queries with different complexity, enabling dynamic control over retrieval depth and pattern~\cite{tang2025mbaragbanditapproachadaptive}. AdaRAG frames the tuning of retrieval ratio and prompt length as bandit convex optimization in edge settings, so that the system can automatically trade off generation quality and end-to-end delay under limited resources~\cite{11044685}. M-RAG further couples multi-agent reinforcement learning with a bandit mechanism that treats database partitions as arms, allowing the system to adaptively select knowledge partitions that best support each query while maintaining scalable retrieval~\cite{wang2024mragreinforcinglargelanguage}. 

A second line of work focuses on bandit-driven knowledge and evidence selection. KnowGPT employs a bandit controller that, conditioned on query context, chooses among candidate knowledge sources and prompt formats, thereby jointly adapting what to retrieve and how to present it to the LLM~\cite{NEURIPS2024_0b8705a6}. In a related direction, Pan et al. use bandit exploration over document subsets to identify which retrieved passages genuinely support the model output, reducing exhaustive subset evaluation while preserving key supporting evidence in RAG pipelines~\cite{pan2025camab}. 

A third direction leverages bandits for multi-objective and hierarchical RAG optimization. For knowledge-graph-driven RAG in non-stationary environments, Tang et al. design a deep multi-objective contextual bandit and employ the Generalized Gini Index to balance accuracy, coverage, and latency, enabling explicit control over competing objectives~\cite{tang2025adapting}. From a systems perspective, AutoRAG-HP adopts a hierarchical MAB framework in which a high-level module selects pipeline variants and a low-level search tunes associated hyperparameters online, providing an automatic mechanism for RAG hyperparameter optimization~\cite{fu2024autoraghpautomaticonlinehyperparameter}.

Despite these advances, existing bandit-based RAG systems often rely on fixed bandit formulations and manually specified reward proxies, which may be brittle when facing new task types, domain shifts, or highly skewed query distributions~\cite{tang2025mbaragbanditapproachadaptive,tang2025adapting}. Moreover, frequent parameter updates and the evaluation of multiple retrieval or configuration strategies can introduce substantial computational overhead, limiting scalability in large-scale or real-time applications. Future work should pursue more adaptive and sample-efficient bandit architectures, richer reward designs grounded in long-term user-centric signals, and tighter coupling between bandit controllers and LLM reasoning to enable robust retrieval--generation coordination across diverse and evolving environments.

\subsection{Inference Optimization}
Inference optimization aims to improve the efficiency and cost-effectiveness of LLM deployment at inference time, encompassing latency reduction, resource allocation, and model or strategy selection under computational constraints~\cite{zhou2024surveyefficientinferencelarge}. In practice, LLM systems must dynamically balance output quality against monetary cost and system load, with these trade-offs varying across queries due to differences in input complexity and user requirements~\cite{xia2023flashllm}. 

These challenges can be naturally framed as sequential decision-making problems, where the system repeatedly selects inference strategies—such as choosing among model variants or allocating computation budgets—and observes feedback only for the executed choice. Bandit algorithms provide a principled abstraction for this setting by enabling adaptive exploration--exploitation trade-offs, allowing inference policies to be learned online under uncertainty.

A first line of work formulates LLM routing and model selection as a bandit problem, treating candidate models or inference backends as arms. Li et al.\ propose a cost-efficient LLM inference framework that adaptively routes queries based on observed performance--cost trade-offs~\cite{li2025llmbanditcostefficientllm}. Panda et al.\ study adaptive LLM routing under explicit budget constraints, framing inference-time model selection as a contextual bandit problem~\cite{panda2025adaptive}. Tongay et al.\ further demonstrate that dynamic model selection via sequential decision-making can effectively respond to changing workloads and deployment conditions~\cite{tongay2025dynamic}. Wei et al.\ introduce a preference-conditioned contextual bandit framework that learns routing policies directly from bandit feedback, enabling fine-grained control over performance--cost trade-offs without full-information supervision~\cite{wei2025learningroutellmsbandit}.

A second line of work focuses on system-level cost-aware inference control, particularly through caching and reuse of LLM outputs. Yang et al.\ model LLM cache selection as a combinatorial bandit with heterogeneous query sizes, showing that adaptive cache management can substantially reduce inference cost while preserving performance guarantees~\cite{yang2025llm}.

Despite these advances, existing bandit-based inference optimization methods often rely on manually designed reward proxies that may be brittle under shifting user preferences or evolving deployment objectives~\cite{li2025llmbanditcostefficientllm,wei2025learningroutellmsbandit}. Moreover, exploration over inference strategies can introduce nontrivial overhead in large-scale or latency-sensitive systems. Future work should investigate more sample-efficient and preference-aware bandit formulations, tighter integration between bandit controllers and LLM reasoning, and extensions to multi-turn and conversational inference settings.

\subsection{Decoding Strategies}
Decoding and reranking components govern how candidate outputs are generated and selected during LLM inference. Modern LLMs often produce multiple candidate responses through stochastic decoding strategies, which are subsequently filtered or reranked based on criteria such as relevance, safety, or user preferences. However, the effectiveness of decoding or reranking strategies can vary significantly across tasks, contexts, and users, and static heuristics often fail to generalize. From a sequential decision-making viewpoint, decoding configuration and candidate selection can be modeled as a bandit problem, enabling adaptive trade-offs between output diversity and quality via online feedback.

Recent work can be grouped into three main lines. 

A first line of work uses bandits for training-free adaptive control of speculative decoding configurations. Hou et al. model alternative speculative setups as arms and apply regret-driven online selection to approach near-oracle configuration choice under changing prefixes and workloads~\cite{hou2025banditspec}. In a more targeted instantiation, Liu et al. cast the choice of draft length as a small-arm bandit and use Thompson sampling to dynamically balance acceleration and verification success~\cite{liu2024speculative}.

A second line shows that some speculative settings provide richer-than-bandit feedback. By leveraging verification structure, Liu et al. demonstrate that one can efficiently score multiple drafters per query, upgrading partial-feedback bandits to (near) full-information online learning and yielding faster identification of strong drafters than standard bandit exploration~\cite{liu2025not}.

A third direction provides theoretical foundations at finer granularity. Shin et al. formalize decoding as irreversible token-by-token selection, establish hardness in general, and identify structural conditions under which sublinear regret becomes achievable, offering a principled lens on when simple decoding rules (e.g., greedy) can be effective~\cite{shin2025tokenized}.

Overall, existing bandit-based decoding approaches are most effective for speed- or acceptance-oriented objectives, but they often rely on hand-designed reward proxies that imperfectly capture user-perceived quality, safety, or long-term utility~\cite{hou2025banditspec,liu2024speculative}. In addition, maintaining online controllers and evaluating multiple decoding configurations can introduce non-negligible control overhead, partially offsetting the gains from acceleration~\cite{liu2025not}. 

Future work should therefore move toward contextual and multi-objective bandit formulations that explicitly incorporate user-centric quality and safety signals, potentially leveraging richer or counterfactual feedback when verification or logging permits~\cite{liu2025not,shin2025tokenized}. Moreover, the bandit framing should be extended beyond decoding acceleration to reranking-specific decisions, such as adaptive best-of-$N$ sizing and reranker selection under delayed or preference-based feedback, which remain largely underexplored in current systems.

\subsection{Adaptation and Personalization}

In adaptation and personalization, the key problem is to align LLM behaviour with user-specific and time-varying preferences while keeping inference and update costs acceptable. Traditional personalization pipelines rely on static fine-tuning or offline clustering, which are slow to adapt, expensive to retrain, and often brittle under heterogeneous feedback across tasks and modalities. Integrating MAB with LLMs offers a principled way to balance exploration of new behaviours and exploitation of learned preferences under limited interaction budgets, especially in conversational agents, recommendation, and summarization systems.~\cite{moerchen2020personalizing,goncc2023user,chen2024online}

Recent work can be grouped into three main lines. 

The first line uses neural bandits to directly personalize soft prompts or representations based on user feedback. Chen et al.~\cite{chen2024online} and Monea et al.~\cite{monea2024llms} model user interactions as contextual bandits, updating soft prompt embeddings to improve the relevance of generated content while preserving a light-weight adaptation mechanism for each user. Moerchen et al.~\cite{moerchen2020personalizing} employ a bandit framework in a speech-based NLU system, using implicit feedback to adjust music recommendation policies without retraining the full model. In parallel, Gönç et al.~\cite{goncc2023user} propose an online learning framework that repeatedly fine-tunes intent classification models with reinforcement-style updates, enabling faster adaptation to novel user inputs in dialog settings and complementing bandit-based approaches. 

A second line focuses on cost-efficient model selection and routing for multi-task personalization. Dai et al.~\cite{dai2024cost} design a compositional MAB that selects among multiple LLMs across tasks, trading off performance against computation by treating model choices as combinatorial arms. Li et al.~\cite{li2025llmbanditcostefficientllm} introduce user-controllable preference vectors that encode the relative importance of accuracy and cost into the bandit reward, and combine them with model identity embeddings to learn routing policies that generalize as user needs and available models change. Wei et al.~\cite{wei2025learningroutellmsbandit} extend this idea to multi-objective bandits, using low-dimensional preference vectors over performance and cost so that model routing can be adjusted online according to user-specified trade-offs at inference time. 

A third line addresses inference-time alignment and adaptive reasoning for personalization. Shin et al.~\cite{shin2025tokenized} reformulate alignment and personalization as a linear bandit problem over tokenized decoding trajectories, combining structured sequence hypotheses with online updates so that the model learns user preferences during inference without further training. Zhang et al.~\cite{zhang2025exploration} compare LLMs and humans in non-stationary environments and show that adjusting reasoning patterns toward more directed exploration improves the adaptability of LLMs, thereby informing bandit-style strategies for personalized decision-making under preference drift.

However, most existing LLM–MAB personalization methods still assume relatively stable or predictable feedback, which may not hold in highly dynamic or sparse interaction regimes. Many approaches rely on soft-prompt adaptation or low-dimensional, manually specified preference vectors, which can miss richer, multi-faceted user objectives and face cold-start issues when only a few generic test items are available.~\cite{chen2024online,monea2024llms,li2025llmbanditcostefficientllm,wei2025learningroutellmsbandit} Future work should explore multi-modal and long-horizon feedback, more expressive preference representations that cover latency, robustness, and risk, and bandit algorithms that can generalize across complex, unstructured environments while updating user models online at inference time.

\section{LLM-Based Enhancements for Bandit Systems}
\label{sec:llm-based_enhancements_for_bandit_systems}

This section systematically examines how large language models (LLMs) enhance various components of multi-armed bandit algorithms. Table~\ref{tab:llm-for-bandit} presents a consolidated overview of representative papers corresponding to each component.

\begin{table}[h]
\caption{LLM-Based Enhancements for Bandit Systems}
\label{tab:llm-for-bandit}

\small
\centering
\setlength{\tabcolsep}{3pt}
\renewcommand{\arraystretch}{1.1}

\begin{tabular}{p{2.8cm} p{5.4cm} p{5.4cm} p{1.8cm}}
\toprule
\textbf{Bandit Component} &
\textbf{Research Problem} &
\textbf{LLM-Enhanced Solutions} &
\textbf{References} \\
\midrule

\multirow{4}{2.8cm}{\raggedright Regret Minimization\\Objective} &
Slow convergence under uniform regret minimization &
LLM-guided information–regret trade-off optimization &
\cite{shufaro2024bits} \\
\addlinespace[1pt]
& Costly exploration in real-time decision scenarios &
Contextual-prior–driven regret minimization &
\cite{xia2024beyond} \\
\addlinespace[1pt]
& Static regret objectives under learning dynamics &
Regret-centric adaptive agent optimization &
\cite{park2024llm} \\
\addlinespace[1pt]
& Regret degradation under shifting rewards &
Dynamic regret objective adjustment &
\cite{de2023llm} \\

\midrule

\multirow{3}{2.8cm}{\raggedright Arm\\Definition} &
Redundant exploration in high-dimensional action spaces &
LLM-based semantic arm compression &
\cite{verma2025neural,liu2018improving} \\
\addlinespace[1pt]
& Static arms under preference drift &
Dynamically redefining or personalizing arms through contextual reasoning &
\cite{xia2024beyond,song2025investigating} \\
\addlinespace[1pt]
& Limited expressiveness of discrete arms &
LLM-generated structured arm manifolds &
\cite{kerim2024multiarmedbanditapproachoptimizing} \\

\midrule

\multirow{3}{2.4cm}{\raggedright Environment} &
Stationary or low-dimensional assumptions fail in non-stationary or semantic-rich settings &
Interpreting evolving contexts to guide adaptation and detect regime shifts. &
\cite{de2023llm,verma2025balancing,alamdari2024jump,salewski2024context} \\
\addlinespace[1pt]
& Realistic environments are difficult to simulate for bandit training &
Generating structured synthetic environments with diverse semantic dynamics &
\cite{kerim2024multiarmedbanditapproachoptimizing,zala2024envgen} \\
\addlinespace[1pt]
& Unstructured and latent environment states &
Language-to-context environment mapping &
\cite{arumugam2025psrl,nessari2025prior} \\

\midrule

\multirow{4}{2.8cm}{\raggedright Reward\\Formulation} &
Noisy and ambiguous feedback &
Context-aware dynamic reward shaping &
\cite{felicioni2024importance,park2024llm} \\
\addlinespace[1pt]
& Sparse or ethically constrained rewards &
LLM-based reward prior and regression &
\cite{sun2025largelanguagemodelenhancedmultiarmed,nessari2025prior} \\
\addlinespace[1pt]
& Misalignment between numeric rewards and human goals &
Natural-language-to-reward translation &
\cite{behari2024decision,verma2025balancing} \\
\addlinespace[1pt]
& Long-horizon credit assignment difficulty &
LLM-mediated preference and algorithmic rewards &
\cite{chen2025greedywins,xia2025incontext} \\

\midrule

\multirow{3}{2.8cm}{\raggedright Sampling\\Strategy} &
Naive exploration is costly in high-dimensional textual action spaces &
LLM-guided informative sampling &
\cite{shufaro2024bits,guo2023towards,chen2024efficient,alamdari2024jump,dai2024cost,sun2025largelanguagemodelenhancedmultiarmed} \\
\addlinespace[1pt]
& Classical sampling rules generalize poorly across heterogeneous tasks &
Learning meta-level exploration policies from textual interaction histories &
\cite{chen2025greedywins,schmied2025greedyllm} \\
\addlinespace[1pt]
& Explicit posteriors are hard to define for language-based rewards &
Using natural language as an implicit posterior to guide exploration &
\cite{hazime2025llm,lim2025textbandit,arumugam2025psrl} \\

\midrule

\multirow{3}{2.8cm}{\raggedright Action\\Decision} &
Poor action ranking under complex context &
LLM-enhanced value and preference estimation &
\cite{baheri2023llms,wang2024llms,guo2023towards,liu2024large,zhang2023using,sun2025largelanguagemodelenhancedmultiarmed} \\
\addlinespace[1pt]
& Rigid numerical decision heuristics &
LLM-as-policy action selection &
\cite{chen2025greedywins,hazime2025llm,lim2025textbandit} \\
\addlinespace[1pt]
& Inconsistent confidence-based decisions &
LLM-aligned posterior-driven action choice &
\cite{schmied2025greedyllm,arumugam2025psrl} \\

\bottomrule
\end{tabular}
\end{table}

\subsection{Regret Minimization Objective}

In MAB problems, the regret minimization objective is to maximize cumulative reward while keeping regret small. Classical algorithms typically assume stationary reward distributions and optimize regret through fixed exploration schedules, which leads to slow convergence and high computational cost in large or rapidly changing environments. Therefore, recent work explores how LLMs can provide prior knowledge, richer contextual understanding, and adaptive feedback so that regret-minimization objectives can be optimized more efficiently in complex decision-making tasks.

Existing studies cluster into several lines of work that reshape the optimization objective by coupling LLM reasoning with bandit exploration–exploitation. 

A first line uses LLMs to guide information-theoretic exploration, explicitly trading off information gain against regret: Shufaro et al.~\cite{shufaro2024bits} use LLM-derived insights to target promising regions of the action space, reducing redundant sampling while maintaining exploration quality, which effectively refines the objective from uniform regret minimization to information-efficient regret control. 

A second line leverages contextual priors and in-context learning to accelerate regret minimization. Xia et al.~\cite{xia2024beyond} show that LLM agents can integrate contextual knowledge about tasks into the bandit objective, allowing the algorithm to down-weight expensive exploration and achieve faster convergence in real-time decision scenarios. 

A third line analyzes regret from the perspective of adaptive LLM agents. Park et al.~\cite{park2024llm} study how LLMs update their behavior based on accumulated experience, providing a regret-centric view of iterative adaptation and suggesting optimization objectives that explicitly encode learning dynamics over time. 

A fourth line focuses on non-stationary environments, where the optimization objective must evolve with the underlying process. De Curt`o et al.~\cite{de2023llm} combine LLM-based real-time insights with bandit strategies to dynamically adjust the effective objective, reducing exploration cost and preserving cumulative reward under shifting reward distributions.

These LLM-enhanced objectives remain constrained by several limitations. Many approaches rely heavily on priors encoded in the LLM that may not generalize across domains, which raises robustness and safety concerns for regret guarantees. Moreover, integrating LLM calls into the optimization loop can introduce substantial computational overhead and latency, limiting scalability in strict real-time settings. Future work should design lightweight surrogates or hybrid architectures in which LLMs shape high-level objectives and priors while core bandit updates remain computationally efficient, and should develop principled methods to balance static LLM priors with online adaptation so that regret remains controlled under abrupt or persistent environment shifts.

\subsection{Arm Definition}
In MAB problems, arms define the set of actions available to the algorithm, and their design is crucial for balancing exploration and exploitation in complex, high-dimensional environments. Traditional methods usually treat arms as discrete indices with limited structure, which makes it difficult to encode heterogeneous contextual variables and semantic relationships, leading to suboptimal action partitions and redundant exploration. Integrating LLMs into arm definition introduces a semantic layer that can jointly capture context, user preferences, and latent structure, thereby enabling more adaptive and interpretable action spaces.

Existing work explores three main lines for LLM-enhanced arm definition. 

First, semantic arm construction and compression uses LLMs to map high-dimensional options and feedback into compact, semantically coherent arm sets. Verma et al.~\cite{verma2025neural} propose a neural dueling bandit framework in which an LLM defines arms and their relations based on semantic similarity, so that the bandit operates over meaningfully clustered actions, improving decision accuracy while reducing the effective action space. Liu et al.~\cite{liu2018improving} show that, in a reinforcement learning setting, a pretrained language model can organize actions into prior-informed categories before interaction, thereby making subsequent online decisions more sample-efficient. 

Second, dynamic and personalized arm redefinition employs LLM reasoning to maintain relevant action sets under non-stationary preferences. Xia et al.~\cite{xia2024beyond} use an LLM within a dueling bandit to continually refine arm definitions according to current contextual cues, so that the compared actions remain aligned with the most informative semantic distinctions in real time. Song et al.~\cite{song2025investigating} study a health incentive task where an LLM generates personalized natural language for each intervention type selected by a contextual MAB, effectively expanding a small set of discrete interventions into a rich manifold of language-level arm realizations tailored to individual users. 

Third, generative arm augmentation combines LLM-based attribute extraction with data generation to define arms as semantically structured data subsets. Kerim et al.~\cite{kerim2024multiarmedbanditapproachoptimizing} use an LLM to automatically extract domain-specific attributes and construct structured prompts for Stable Diffusion, and then treat the resulting synthetic data clusters, partitioned by semantic attributes, as distinct arms for downstream sample selection.

These approaches demonstrate that LLM-guided arm definition can substantially improve the adaptability and robustness of MAB algorithms in high-dimensional, context-sensitive environments. However, they also introduce additional computational overhead and latency, which can limit deployment in strict real-time settings and raise concerns about the stability of LLM-generated semantics. Future research should focus on more efficient pipelines for LLM-based arm construction, scalable personalization that balances individual feedback with generalizable arm structures, and theoretical analyses that clarify how semantic arm design affects regret and robustness in large-scale bandit systems.

\subsection{Environment}

In MAB settings, the environment determines how contexts, rewards, and latent dynamics evolve, so non-stationarity and adversarial shifts can rapidly invalidate fixed exploration–exploitation strategies.  Previously proposed reinforcement learning (RL) and Bandit algorithms~\cite{sutton2018reinforcement,lattimore2020bandit} are mainly suited for stationary environments.  Classical methods usually assume stationary reward distributions or simple drift models and rely on low-dimensional statistics, which makes them brittle under high-dimensional semantic feedback, delayed effects, or abrupt regime changes. Reference~\cite{Padakandla2021banditenvironments} summarizes the solutions of  RL algorithms in both finite and infinite horizons for non-stationary environments.
Integrating LLMs into the environment component is therefore valuable, as their contextual and sequence reasoning enables richer state representations, change detection, and forecasting of future conditions in which bandit policies will operate.

Existing work can be grouped into three main lines. 

First, several studies use LLMs to guide adaptation in non-stationary or adversarial environments. These approaches let the LLM parse evolving contexts and histories to adjust bandit beliefs, anticipate regime shifts, or regularize decisions against adversarial manipulation. For example, De Curtò et al.~\cite{de2023llm} design an LLM-informed policy for non-stationary settings that conditions arm selection on changing contextual patterns;  Verma et al.~\cite{verma2025balancing}  design a social-choice–guided framework in which LLMs generate candidate reward functions and an adjudicator selects those best aligned with multi-group preferences in structured RMAB environments; Alamdari et al.~\cite{alamdari2024jump} use LLM-generated priors to warm-start policies in new environments and reduce exploration time; and Salewski et al.~\cite{salewski2024context} show that LLMs can track dynamic contexts in real time, identify systematic biases, and refine policy updates accordingly. 

Second, a complementary line uses LLMs to construct more structured environments for training bandit agents. Kerim et al.~\cite{kerim2024multiarmedbanditapproachoptimizing} employ LLMs to analyse the image domain and define an organized attribute space in which diffusion models generate diverse yet semantically coherent samples, enabling the bandit to explore a structured synthetic dataset and select the most informative image subsets. Similarly, Zala et al.~\cite{zala2024envgen} use LLMs to generate and adapt synthetic environments that mimic complex real-world dynamics, thereby improving agent training and long-term adaptability. 

Third, several works exploit LLMs to transform unstructured narratives into bandit-ready contextual representations, effectively enriching the observable environment. Arumugam et al.~\cite{arumugam2025psrl} summarize interaction trajectories in natural language and use these summaries to update posterior beliefs over rewards and transitions in a Bayesian fashion without explicit parametric modeling; and Nessari et al.~\cite{nessari2025prior} map unstructured clinical notes into structured patient features that serve as high-dimensional, interpretable contexts for online treatment recommendation in real-world medical environments.

Despite these advances, most LLM-enhanced environment models still rely on simplified assumptions about how drift, adversaries, and feedback loops arise, which limits their robustness in truly open-world settings. Future work should more tightly couple LLM-generated priors with on-line adaptive learning so that policies remain flexible in dynamic environments without overfitting to handcrafted descriptions or pre-specified knowledge. It is also important to develop models that generalize across heterogeneous tasks and environment types, proactively predict harmful shifts, improve robustness in adversarial regimes, and leverage LLM-generated synthetic environments to expose agents to the diversity and hidden biases of real-world conditions while maintaining fairness in downstream decisions.

\subsection{Reward Formulation}
Reward formulation is a central but challenging component in MAB, because the agent only observes partial, noisy, and often delayed feedback while needing to optimize long-term performance. Traditional designs rely on fixed numerical rewards or manually engineered transformations, which struggle with non-stationary preferences, multi-objective trade-offs, and ambiguous or language-based feedback. LLMs provide a way to translate rich contextual and linguistic signals into structured reward objects, thereby aligning MAB objectives more closely with high-level human goals and complex environments.

Existing work can be grouped into four main lines. 

First, a set of studies uses LLMs for dynamic and context-aware reward shaping in non-stationary or multi-agent settings.  Felicioni et al.~\cite{felicioni2024importance}, and Park et al.~\cite{park2024llm} further show how LLMs can detect changes in reward patterns, disambiguate noisy or underspecified feedback, and adapt online reward structures to improve cumulative performance. 

Second, some work treats the LLM as a reward surrogate or prior model. Sun et al.~\cite{sun2025largelanguagemodelenhancedmultiarmed} propose TS-LLM and RO-LLM frameworks in which the LLM performs in-context regression from contextual descriptions to expected arm returns, replacing classical parametric regressors within Thompson sampling and robust optimization to better handle nonlinear reward surfaces. Nessari et al.~\cite{nessari2025prior} use LLM-generated structured features to train counterfactual outcome models in healthcare, and then use the predicted potential treatment effects as prior rewards for the bandit, which mitigates sparse and ethically constrained reward observations. 

Third, LLMs are used to construct reward functions directly from natural language specifications of objectives and preferences. Behari et al.~\cite{behari2024decision} let an LLM convert high-level public-health guidelines into explicit RMAB reward functions and iteratively refine them, so that the decision process tracks evolving human policy priorities. Verma et al.~\cite{verma2025balancing} extend this idea by generating multi-objective reward candidates from prompts and applying a social-choice-style evaluation step to select reward structures that best reconcile conflicting subgroup priorities. 

Fourth, several studies integrate algorithmic and preference-based signals into the LLM’s reward modeling. Chen et al.~\cite{chen2025greedywins} design strategic rewards based on instantaneous regret and consistency with UCB-style decisions, which are then processed by an LLM policy optimized via RL to improve exploration and alleviate long-horizon credit assignment. Xia et al.~\cite{xia2025incontext} propose a dueling-bandit setting where the LLM transforms pairwise preference feedback expressed in natural language into decision signals, effectively replacing explicit numerical rewards and improving decision efficiency in preference-driven tasks.

Despite these advances, LLM-based reward formulation still faces several limitations. Current methods often depend on costly model queries and may inherit biases or inconsistencies present in language feedback, which can distort long-run bandit behavior. Future work should develop more sample-efficient, interpretable, and theoretically grounded reward-learning pipelines that combine LLM priors with off-policy bandit data, and systematically study robustness under distribution shift, multi-objective trade-offs, and misaligned or strategic human feedback.

\subsection{Sampling Strategy}

In MAB problems, the sampling strategy defines how actions are selected based on current uncertainty estimates, governing the balance between exploration and exploitation throughout the learning process. LLM-enhanced sampling strategies aim to address this classical dilemma when actions, context, and feedback are expressed in high-dimensional text. In such settings, naive exploration is costly and hand-crafted rules based only on numeric statistics struggle to capture semantic structure, shifting user preferences, or heterogeneous costs. Integrating LLMs into the sampling module allows the algorithm to use textual histories and prior knowledge to propose more informative arms, potentially improving sample efficiency under tight interaction or computation budgets.

Existing studies can be broadly grouped into three methodological directions.

A first line uses LLMs to guide information-aware and cost-aware exploration on top of standard bandit statistics. Shufaro et al.~\cite{shufaro2024bits} formalize a trade-off between regret and information gain, and query an LLM to identify contextually informative arms so that exploration reduces redundant sampling while controlling cumulative regret. Related work uses LLM reasoning to prioritize arms in high-dimensional contextual spaces or to adapt exploration intensity over time, for example by ranking candidate arms using textual cues~\cite{guo2023towards}, adjusting exploration strength based on observed feedback~\cite{chen2024efficient}, or “jump-starting’’ learning with LLM proposals that reduce early-stage exploration~\cite{alamdari2024jump}. Dai et al.~\cite{dai2024cost} study online selection among multiple LLMs, using LLM-provided contextual insight to design cost-efficient sampling policies under resource constraints, while Sun et al.~\cite{sun2025largelanguagemodelenhancedmultiarmed} tune the generation temperature of an LLM to implement a Thompson-style randomized sampling mechanism, treating controlled output stochasticity as a proxy for posterior uncertainty without explicitly maintaining a parametric posterior. 

A second line lets LLMs directly learn and represent exploration policies that behave like meta-bandits. Chen et al.~\cite{chen2025greedywins} train LLMs via supervised fine-tuning or reinforcement learning to imitate classical strategies such as UCB, so that the model maps textual reward histories into explicit next-arm choices and can generalize exploration behavior across different reward distributions. Schmied et al.~\cite{schmied2025greedyllm} further enhance self-generated chain-of-thought, combining it with mechanisms such as $\epsilon$-greedy selection, self-consistency, and reward-gain heuristics so that the LLM’s reasoning trace actively shapes when to explore or exploit. 

A third direction leverages language as an implicit posterior over rewards or environments, using natural-language inference in place of explicit probabilistic sampling. Hazime et al.~\cite{hazime2025llm} ask the LLM to decide in natural language when to switch arms based on textual reward histories, thereby approximating exploration–exploitation trade-offs without explicit numeric confidence bounds. Lim et al.~\cite{lim2025textbandit} treat text-based success and failure feedback as updates to the LLM’s subjective reward estimates, effectively using language as the sampling basis instead of formal posterior sampling. Extending this idea to reinforcement learning, Arumugam et al.~\cite{arumugam2025psrl} prompt LLMs to generate language-level posterior hypotheses about the environment and then sample "plausible MDPs" from these hypotheses, realizing a PSRL-like exploration mechanism where uncertainty is encoded and manipulated in natural language before driving action selection.

Despite these advances, current LLM-based sampling strategies often rely on heuristic prompt designs and narrow experimental settings, which limits theoretical understanding and robustness in large, noisy, or highly multi-armed environments. For example, purely language-driven exploration can be sensitive to decoding parameters such as temperature, leading to unstable sampling quality when rewards are noisy or the arm set is large~\cite{hazime2025llm}. Future work should therefore seek stronger theoretical guarantees for language-mediated exploration, broader empirical validation across dynamic and non-stationary tasks, and more resource-efficient designs that reduce LLM call overhead. Promising directions include post-learning schemes that feed back long-horizon bandit traces into the LLM, and tighter integration with external memory, tool use, and adaptive prompting so that LLM-guided sampling becomes more systematic, controllable, and interpretable.

\subsection{Action Decision}

The action decision component determines which arm is selected given current statistics and context, and is therefore central to maximizing long-term reward in dynamic environments. Classical policies such as greedy rules, upper confidence bounds, or probability matching rely on hand-crafted numerical heuristics that struggle with high-dimensional context, non-stationary preferences, and semantically rich feedback. Integrating LLMs with MAB aims to inject prior knowledge and contextual reasoning into this decision process, enabling more adaptive and interpretable action choices.

Recent work on LLM-enhanced action decision can be grouped into three main directions. 

First, several studies use LLMs to refine value and preference estimates before applying a bandit rule. Baheri et al.~\cite{baheri2023llms} propose an LLM-augmented contextual bandit framework in which the LLM encodes textual contexts into dense semantic representations to improve action selection. Wang et al.~\cite{wang2024llms} exploit LLM-based semantic analysis of context and interaction history to identify actions with potentially higher reward in dynamic user-interest exploration. Guo et al.~\cite{guo2023towards} combine LLM priors with high-dimensional action spaces to guide arm selection, while Liu et al.~\cite{liu2024large} treat the LLM as an evolutionary optimizer that improves policy quality with reduced exploration. Zhang et al.~\cite{zhang2023using} let an LLM continuously tune hyperparameters in real-time tasks so that updated configurations induce better downstream decisions, and Sun et al.~\cite{sun2025largelanguagemodelenhancedmultiarmed} extend LLM prediction to pairwise preference comparison, embedding estimated win probabilities into a dueling bandit to obtain more stable preference-based action rankings. 

Second, another line of work treats the LLM itself as the decision rule that maps summaries of interaction data directly to arms. Chen et al.~\cite{chen2025greedywins} feed statistical aggregates such as empirical means or confidence summaries into an LLM, which learns decision heuristics resembling greedy or UCB-style rules and assumes full responsibility for action choice. Hazime et al.~\cite{hazime2025llm} and Lim et al.~\cite{lim2025textbandit} instead prompt the LLM with natural-language interaction histories so that actions are selected through textual reasoning rather than explicit bandit formulas, effectively replacing classical decision policies by language-based ones. 

Third, some methods explicitly align LLM reasoning with posterior or confidence-based decision mechanisms. Schmied et al.~\cite{schmied2025greedyllm} use reinforcement-learning fine-tuning to enforce consistency between the LLM’s inferred UCB estimates and its chosen actions, mitigating greedy bias and “knowing-but-not-doing’’ behavior, while Arumugam et al.~\cite{arumugam2025psrl} provide sampled hypothetical MDPs to a decision LLM so that it reasons in language space about the optimal action under posterior samples, thereby realizing a PSRL-style decision flow via LLM inference.

Despite these advances, current approaches rarely treat the bandit action decision module itself as an object of systematic optimization with strong theoretical guarantees, and many works implicitly focus on improving exploration rather than jointly shaping exploration and exploitation. Empirical studies also reveal instability in language-driven decision rules: for example, Hazime et al.~\cite{hazime2025llm} observe random drift and sensitivity to complex reward structures, which can cause local fluctuations to be misinterpreted as long-term trends. Future research may thus combine structured history summarization, constrained decision templates, and external evaluators with chain-of-thought control, restrictive policies, or reward models, aiming to obtain more stable, theory-grounded LLM-based action decision mechanisms that balance semantic reasoning with reliable bandit behavior.

\subsection{Evaluation Issues}
The reviewed papers apply LLM-enhanced bandit algorithms using both synthetic datasets, such as non-stationary and restless bandit simulations, and real-world datasets like MovieLens and Yahoo! Front Page. Evaluation metrics focus on cumulative reward, regret, and precision-based metrics like Precision@k (P@k) and click-through rate (CTR), commonly used in recommendation systems. LLMs also contribute to reducing exploration costs, with metrics like time to convergence and computational efficiency being critical in multi-task and real-time applications. Table~\ref{tab:datasets} and ~\ref{tab:metrics} provides representative examples of datasets and evaluation metrics commonly adopted in prior work, and is not intended to be exhaustive.

\begin{table}[h]
\caption{Representative experimental datasets used in LLM-enhanced bandit studies}
\label{tab:datasets}
\centering
\small
\begin{tabular}{l l}
\toprule
\textbf{Dataset Type} & \textbf{Dataset (Reference)} \\
\midrule
Synthetic &
Non-stationary bandit simulations~\cite{de2023llm} \\

& Restless bandit simulations~\cite{zhao2024towards,verma2025balancing} \\

& Contextual bandit simulations~\cite{baheri2023llms} \\

\addlinespace[3pt]
Real-world &
MovieLens~\cite{wang2024llms} \\

& Yahoo! Front Page~\cite{kiyohara2024prompt} \\

& OpenAI Gym~\cite{liu2024large} \\

& Amazon Product Data~\cite{guo2023towards} \\
\bottomrule
\end{tabular}
\end{table}

\begin{table}[t]
\caption{Representative evaluation metrics adopted in LLM-enhanced bandit studies}
\label{tab:metrics}
\centering
\small
\begin{tabular}{ll}
\toprule
\textbf{Metric} & \textbf{Representative Studies} \\
\midrule
Cumulative reward &
Shufaro et al.~\cite{shufaro2024bits}; 
Sun et al.~\cite{sun2025largelanguagemodelenhancedmultiarmed} \\
\addlinespace[2pt]

Regret &
Xia et al.~\cite{xia2024beyond,xia2025incontext}; 
Park et al.~\cite{park2024llm} \\
\addlinespace[2pt]

Precision@k (P@k) &
Wang et al.~\cite{wang2024llms}; 
Baheri et al.~\cite{baheri2023llms} \\
\addlinespace[2pt]

Click-through rate (CTR) &
Kiyohara et al.~\cite{kiyohara2024prompt} \\
\addlinespace[2pt]

Time to convergence &
Alamdari et al.~\cite{alamdari2024jump} \\
\addlinespace[2pt]

Exploration cost &
Guo et al.~\cite{guo2023towards}; 
Chen et al.~\cite{chen2024efficient} \\
\addlinespace[2pt]

Computational efficiency &
Dai et al.~\cite{dai2024cost} \\
\addlinespace[2pt]

Prompt--utility ratio &
Verma et al.~\cite{verma2025balancing} \\
\bottomrule
\end{tabular}
\end{table}

Only a small number of papers in LLM-enhanced bandit research focus on providing strong theoretical guarantees for regret minimization, particularly in non-stationary and adversarial settings. These works aim to demonstrate how LLMs can improve exploration-exploitation trade-offs, with papers like Shufaro et al. (2024)~\cite{shufaro2024bits} and Xia et al. (2024)~\cite{xia2024beyond} offering rigorous regret upper bounds. For example, Shufaro et al. (2024)~\cite{shufaro2024bits} analyze the trade-off between regret and information gain, and Xia et al. (2024)~\cite{xia2024beyond} prove regret bounds in dueling bandit settings with LLM agents. However, the theoretical regret proofs often rely on simplifying assumptions about the environment or the behavior of the LLMs themselves. For instance, Alamdari et al. (2024)~\cite{alamdari2024jump} leverage LLM-generated prior knowledge to reduce exploration cost, but their regret bounds depend heavily on the quality and accuracy of the priors, which might not always hold in real-world applications.

A major limitation is the lack of formal guarantees for LLM performance in complex spaces, making regret proofs dependent on strong assumptions like well-structured environments. Without such conditions, proving regret bounds for these algorithms is challenging. Given these constraints, future research should prioritize practical effectiveness in real-world applications over theoretical proofs. Researchers need to develop methods that perform well in dynamic environments, even without strong theoretical guarantees, striking a balance between theoretical rigor and practical validation.

\section{Challenges and Future Opportunities}
\label{sec:challenges_and_future_opportunities}

Bandit algorithms and Large Language Models (LLMs) exhibit a mutually reinforcing relationship. On one hand, bandits offer principled frameworks for optimizing LLM training, inference, and adaptation in dynamic environments. On the other hand, LLMs provide rich contextual reasoning and prior knowledge that can significantly enhance the design of next-generation bandit algorithms. This section outlines future opportunities along both directions, beginning with \textit{Bandits for LLMs} and followed by \textit{LLMs for Bandits}.

\paragraph{Bandits for LLMs.}
Current research demonstrates that bandit algorithms are indispensable for enhancing the efficiency, adaptability, and robustness of LLM systems. However, there are several challenges that need to be addressed in order to fully leverage bandit algorithms in LLM systems:

\begin{itemize}
    \item \textbf{Exploration-Exploitation Balance in High-Dimensional Decision Spaces:} LLMs generate outputs in complex, high-dimensional decision spaces (e.g., word choice, sentence structure, context adaptation). Traditional bandit algorithms often struggle to efficiently explore such spaces. Developing methods to balance exploration and exploitation in these high-dimensional spaces remains a key challenge.
    
    \item \textbf{Sparse and Noisy Feedback Signals:} The feedback from LLMs is often sparse, noisy, and difficult to quantify (e.g., user satisfaction, output relevance). Converting these unstructured feedback signals into actionable reward signals for bandit algorithms is a major obstacle.
    
    \item \textbf{Long-Term Reward Prediction and Adaptation:} Many LLM tasks, such as multi-turn dialogue or multi-step text generation, involve long-term dependencies. Bandit algorithms are typically optimized for short-term rewards, making it difficult to account for long-term rewards and adapt to them. Developing methods to predict and optimize long-term rewards remains a significant challenge.
    
    \item \textbf{Non-Stationary Environments:} LLMs, particularly in dialogue systems or multi-task learning, are highly sensitive to contextual changes. The rewards in these environments are often non-stationary, and traditional bandit algorithms are not designed to handle such variability effectively. Bandit algorithms need to be adapted to handle non-stationary feedback.
    
    \item \textbf{Multi-Task and Multi-Objective Optimization:} LLM systems often need to balance multiple tasks and objectives, such as accuracy, diversity, and user satisfaction. Bandit algorithms must be extended to handle these complex trade-offs and efficiently allocate resources across multiple tasks or objectives.
\end{itemize}

Several future directions can further bridge theory and practice in using bandit algorithms to enhance LLMs:

\begin{itemize}
    \item \textbf{Continual Learning Optimization:} Use bandits to dynamically manage continual learning in LLMs, selecting the most informative data for long-term performance. This would ensure that LLMs can efficiently adapt to new data without catastrophic forgetting.
    
    \item \textbf{Automatic Prompt Engineering:} Develop bandit-driven systems that automate prompt design, optimizing LLM performance across different tasks. This would enable adaptive prompting, improving the overall efficiency and accuracy of LLM-based systems.
    
    \item \textbf{Multi-Task and Multi-Objective Learning Bandits:} Develop multi-task and multi-objective bandits that allocate resources to optimize LLM performance across several tasks and objectives simultaneously, balancing the trade-offs between different goals.
    
    \item \textbf{LLM Compression and Efficiency:} Apply bandits to intelligently explore model compression and pruning strategies, improving computational efficiency. Bandit algorithms can help in selecting the most efficient model configurations while maintaining high performance.
\end{itemize}

\paragraph{LLMs for Bandits.}
LLMs are increasingly powerful at understanding context, generating insightful priors, and making decisions in dynamic environments. These capabilities offer significant potential for enhancing bandit algorithms, particularly in handling complex environments with high-dimensional data. Several challenges and research opportunities exist in leveraging LLMs to advance bandit algorithms:

\begin{itemize}
    \item \textbf{Adaptive Exploration with Continuous Learning:} Bandit algorithms must continuously adapt to changes in the environment. LLMs can help improve exploration-exploitation strategies by learning from feedback in real time and adjusting exploration behaviors over time.
    
    \item \textbf{LLM-Driven Multi-Modal Bandits:} Traditional bandit algorithms often operate in single-modal contexts (e.g., text or numerical data). LLMs, with their ability to process multiple types of data (e.g., text, images), can be integrated into multi-modal bandit systems, enabling more sophisticated decision-making environments that require integrating different data modalities.
    
    \item \textbf{Human-in-the-Loop Bandits:} Incorporating real-time human feedback into bandit algorithms can greatly enhance their performance in adaptive settings. LLMs can act as mediators between the system and the human input, guiding decision-making based on the user's preferences or contextual changes.
    
    \item \textbf{Regret-Aware Bandit Systems:} LLMs can predict long-term outcomes and adjust bandit strategies accordingly, minimizing regret in non-stationary and adversarial environments. This ability to incorporate sophisticated reasoning could improve decision-making under uncertainty.
    
    \item \textbf{Contextual Bandits for Dynamic Decision-Making:} LLMs can leverage their contextual understanding to make more informed decisions in dynamic, uncertain environments. Developing contextual bandits that use LLMs to adaptively select actions based on evolving inputs (e.g., user interactions, environmental changes) will significantly enhance the flexibility and robustness of decision-making systems.
\end{itemize}

As bandit algorithms become increasingly integrated with large language models, developing rigorous theoretical guarantees will become substantially more challenging due to the high complexity, non-linearity, and implicit reasoning structures of LLMs. Nevertheless, future research may shift toward evaluating such hybrid bandit--LLM systems primarily through their empirical performance in real-world applications, rather than relying solely on traditional theoretical analyses. This shift could drive the practical deployment of these systems in a variety of complex, real-world environments.

\section{Conclusion}
\label{sec:conclusion}

To the best of our knowledge, this work represents the first systematic review that analyzes the LLM–MAB intersection from a component-based and algorithmic perspective. Through a systematic breakdown of key components, we demonstrate how LLMs enhance the performance of bandit algorithms and how bandit frameworks, in turn, optimize various LLM tasks. This survey establishes a solid foundation for future research, highlighting the importance of application-driven studies that strike a balance between theoretical rigor and practical impact. We hope this work will inspire and guide future advancements in this rapidly evolving field.

\bibliographystyle{unsrt}  
\bibliography{references}  

\end{document}